\documentclass[10pt,twocolumn,letterpaper,final]{article}

\usepackage{cvpr}
\usepackage{booktabs}
\usepackage{times}
\usepackage{epsfig}
\usepackage{graphicx}
\usepackage{amsmath}
\usepackage{amssymb}
\usepackage{makecell}
\usepackage{subcaption}
\usepackage{pdfpages}

\usepackage[pagebackref=true,breaklinks=true,letterpaper=true,colorlinks,bookmarks=false]{hyperref}

\cvprfinalcopy 

\def\cvprPaperID{3999} 

\newcommand{\ignore}[1]{}

\begin{document}

\title{In the Wild Human Pose Estimation Using \\Explicit 2D Features and Intermediate 3D Representations}

\author{Ikhsanul Habibie, Weipeng Xu, Dushyant Mehta, Gerard Pons-Moll, Christian Theobalt\\
Max Planck Institute for Informatics, Saarland Informatics Campus, Saarbrucken, Germany\\
{\tt\small \{ihabibie, wxu, dmehta, gpons, theobalt\}@mpi-inf.mpg.org}
}

\maketitle

\begin{abstract}
Convolutional Neural Network based approaches for monocular 3D human pose estimation 
usually require a large amount of training images with 3D pose annotations. While it is feasible to provide 2D joint annotations for large corpora of in-the-wild images with humans, providing accurate 3D annotations to such in-the-wild corpora is hardly feasible in practice. Most existing 3D labelled data sets are either synthetically created or feature in-studio images. 3D pose estimation algorithms trained on such data often have limited ability to generalize to real world scene diversity. We therefore propose a new deep learning based method for monocular 3D human pose estimation that shows high accuracy and generalizes better to in-the-wild scenes. It has a network architecture that comprises a new disentangled hidden space encoding of explicit 2D and 3D features, and uses supervision by a new learned projection model from predicted 3D pose. Our algorithm can be jointly trained on image data with 3D labels and image data with only 2D labels. It achieves state-of-the-art accuracy on challenging in-the-wild data.
\end{abstract}


\section{Introduction}

Human motion capture has a wide range of applications in computer animation and also other areas such as biomechanics, medicine, and human-computer interaction. However, the standard 3D human motion capture systems typically require marker suits and/or multiple cameras recording in a controlled setting which are expensive and complicated to set up, and are impractical outside of the lab or studio environments. Methods that infer 3D pose only from monocular images overcome many such limitations and make 3D pose estimation more widely applicable. However, due to the under-constrained nature of monocular 3D pose estimation, achieving accurate 3D prediction is still a challenging task.

Recent progress of Convolutional Neural Networks (CNN) \cite{Krizhevsky2012imagenet} has enabled promising learning-based methods for 3D human pose estimation from a single color image. Training such methods typically requires a large amount of RGB images annotated with reference 3D poses from either marker-based or markerless multi-camera motion capture systems \cite{Stoll_2011, EEJTP15, rhodin2016general, joo2018total}, synthetic data \cite{Chen2016synthesizing}, or IMU-based systems \cite{DIP:SIGGRAPHAsia:2018,SIP2017,vonMarcard2018}. Owing to this complex reference data capturing, diversity in real world appearance or pose is hard to achieve in training data, which limits the generalization of trained networks on in-the-wild scenes.

\begin{figure}
\centering
\includegraphics[width=1\linewidth]{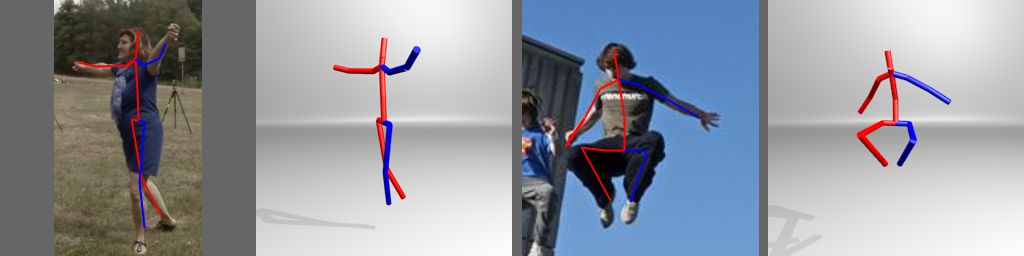}
\label{fig:teaser}
\vspace{-0.3cm}
\caption{3D pose prediction using our method for general scenes. Please refer to Sec. \ref{approach} for the details of our method and Sec.~\ref{results} for results and evaluations.}
\end{figure}

To improve in-the-wild generalization, previous work has leveraged features learned on in-the-wild annotated 2D pose data. Some methods \cite{mono-3dhp2017, mehta2017vnect} proposed to finetune this learned representation on 3D pose prediction using 3D pose datasets captured in a studio. Others \cite{Zhou_2017_ICCV} use this learned representation as an initialization to jointly predict both 2D key points and depth information. For images where the 3D annotations are available, both 2D keypoints and depth are supervised, with supervision coming from geometric constraints otherwise.
In this way, networks carry over features useful for in-the-wild 2D for better 3D pose estimation in out-of-studio settings.

Using a strong pre-existing pose prior, like a parametric body model, can also help a network to predict more accurate 3D poses if labelled 3D training data is scarce~\cite{Yang_3dposeCVPR2018, kanazawaHMR18}.

Since 3D pose labels on general scene images are hard to obtain while larger annotated 2D training corpora exist, several deep learning based methods resort to using 2D pose as the target prediction, followed by an additional 3D pose lifting step \cite{deepercut2016,wang2014robust,yasin2016a,Bogo:ECCV:2016,tome2017lifting,chen_2017_3d,JMartinez:ICCV:2017}. 
Using such, \cite{JMartinez:ICCV:2017} showed that 2D pose data alone is enough to train a network that achieves promising 3D pose estimation accuracy. 
However, solely predicting 3D from 2D pose is an inherently ambiguous task and in these approaches important 3D pose cues from the image are neglected.

In this paper, we introduce a new convolutional neural network architecture for 3D pose estimation that achieves state-of-the-art accuracy on challenging in-the-wild data. It introduces two main innovations that enable us to effectively train the network using both, more scarcely available image data with 3D annotation and more easy to generate image data with only 2D annotation. 

The first innovation is inspired by 2D-to-3D pose lifting~\cite{JMartinez:ICCV:2017}, but maintains the network's capability to explicitly utilize 3D cues in images. 
To this end, we design some channels of the convolutional latent space to encode explicit 2D keypoint features in heatmaps, leaving the rest of the features to contain ``depth'' information about the human pose. 
Separating the 2D and depth, and supervising 2D with additional in-the-wild data, which has been the primary driver of accurate 2D pose estimation methods \cite{Wei2016ConvolutionalPM, Newell2016StackedHN, cao2017realtime}, allows the network to consequently predict 3D pose more reliably even under a significant shift of the input appearance between the training and testing time. These 2D pose features can be trained jointly with depth features on data with 3D annotations, or trained 
independently on data with 2D annotations, while in both cases improving overall network performance. 

The second innovation is a supervision approach that reduces 3D-to-2D ambiguity when training on data with 2D annotations only. To this end, we design a neural network that learns how to estimate the location of 2D body joints by using the 3D human pose predicted from the earlier network layers as latent features. More specifically, we learn to predict the weak perspective camera parameters of the given monocular image input that project the predicted 3D pose to the 2D space. During training, this projection loss can be used to update the information of 3D joint positions regardless if the training image has 3D labels or only 2D labels.

Our approach achieves a state-of-the-art accuracy of 70.4\% 3D PCK on the MPI-INF-3DHP benchmark with challenging outdoor scenes, even when trained only using images with 3D pose labels from the H3.6M \cite{h36m_pami} studio dataset. When jointly training on larger corpora of in-studio images with 3D labels and in-the-wild data with 2D labels, we achieve 91.3\% 3D PCK on MPI-INF-3DHP which outperforms all previous methods. 

\begin{figure*}
  \centering
  \includegraphics[width=0.85\textwidth]{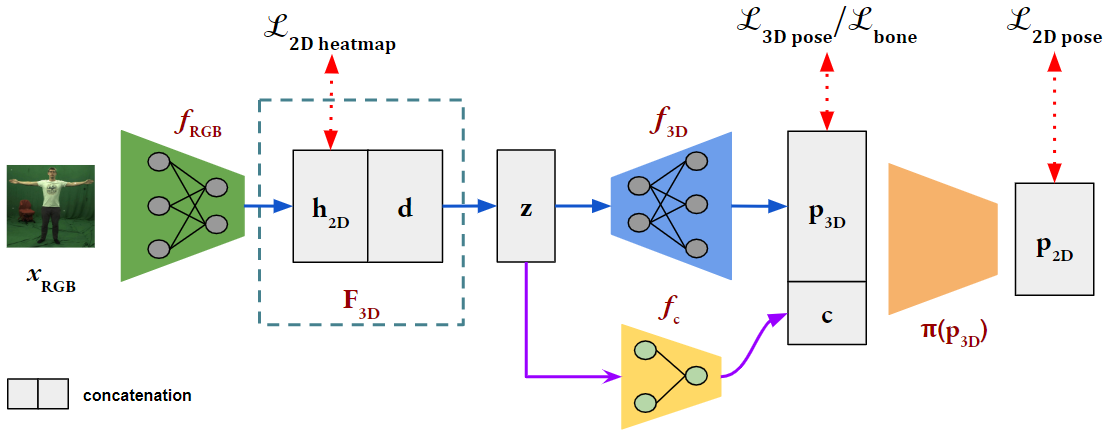}
  \caption{The overview of our proposed architecture. We use a CNN $f_{RGB}$ to learn 3D pose features represented as 2D heatmap locations $\mathbf{h}_{2D}$ and additional 3D pose cues $\mathbf{d}$ in the latent space. Both information are used to predict a root centered 3D pose $\mathbf{p}_{3D}$ and viewpoint parameters $\mathbf{c}$ using networks $f_{3D}$ and $f_{c}$, respectively. Finally, we concatenate $\mathbf{p}_{3D}$ and $\mathbf{c}$ to learn 2D keypoint information $\mathbf{h}_{2D}$, allowing the network to update 3D pose information even if 3D labels are not available.}
  \vspace{-0.3cm}
  \label{fig:architecture}
\end{figure*}

\section{Related Work} \label{relatedwork}

Human pose estimation is an actively studied area in computer vision. We focus our discussion on recent learning-based approaches that are relevant to our work.

\noindent
\textbf{3D pose from 2D keypoint detection.} Due to the robustness of some recent CNN-based 2D pose detection methods \cite{toshev2014deep,Tompson2015EfficientOL,Wei2016ConvolutionalPM, Newell2016StackedHN,cao2017realtime}, many 3D pose estimation method reformulate the task as a combination of 2D keypoints prediction and body depth regression. Mehta \etal \cite{mehta2017vnect} combine 2D heatmap prediction with 3D location maps to estimate the position of each joint in the 3D space. Zhou \etal \cite{Zhou_2017_ICCV} propose a weak supervision training scheme using a stacked hourglass network \cite{Newell2016StackedHN} on both in-the-wild 2D data and studio data with 3D labels. The network is trained to predict 2D pose on both studio and outdoor dataset and at the same time also learns to predict depth information from the 3D labeled data. Yang \etal \cite{Yang_3dposeCVPR2018} also use similar weak supervision, but they extend this idea by introducing an adversarial network that learns how to differentiate between a ground truth and a predicted pose generated by the 3D pose prediction network. Another similar line of work is proposed by Dabral \etal \cite{Dabral:ECCV:2018} which improves this approach further by using body symmetry constraints and a separate temporal prediction network to achieve better 3D prediction stability across sequential frames. 

To take the full advantage of the detection-based method, Pavlakos \etal \cite{pavlakos2017volumetric} proposed using a volumetric representation as an extension of the 2D joint heatmaps in the 3D space. However, this formulation is computationally expensive to perform even after using their coarse-to-fine strategy proposed to mitigate this issue. 

\noindent
\textbf{Direct 3D pose prediction.} Instead of using the combination of 2D and depth prediction, several works regress 3D body keypoints directly. Tekin \etal \cite{Tekin2016StructuredPO} enhance a direct 3D prediction network by learning human body structure using a pose autoencoder. 

Mehta \etal \cite{mono-3dhp2017} use multiple intermediate supervision tasks, such as predicting the output at multiple network levels and predicting 2D heatmaps as an additional objective. They use two step training approach to improve generalization. 
The network is firstly trained to learn 2D joint heatmaps and then refined on the task of directly predicting 3D joint location maps from 3D annotated studio data. Instead of directly predicting the keypoints, \cite{zhou2016deep} regresses the joint angles on a kinematic body model, assuming that the bonelength of the subject is known. Sun \etal \cite{Sun:ECCV:2017} use a geometry aware formulation that also predicts bone length and bone vector orientation instead of only regressing 3D keypoint locations.

Rhodin \etal \cite{rhodin2018learning} proposed a multi-view consistent prediction approach during training to refine neural network's monocular pose prediction on general scenes. 
But it requires synchronized multi-camera footage to train. Multi-view settings can also be used to perform unsupervised or semisupervised learning on human pose estimation by training the network to learn a geometry aware latent space that can generate novel view on different cameras \cite{rhodin2018unsupervised}.

\noindent
\textbf{3D lifting without depth information.} 
Some methods compute 3D pose by estimating the depth from the detected 2D keypoints only. Tome \etal \cite{tome2017lifting} performs a sequence of 3D lifting and reprojection to iteratively improve prediction quality. Chen \etal \cite{chen_2017_3d}
find a closest 3D pose from a library of human poses that best matches the detected 2D pose. \cite{JMartinez:ICCV:2017} use a fully connected neural network with residual connection can achieve accurate 3D pose estimation performance using 2D ground truth or a very accurate 2D keypoint detection as input. Regardless, these approaches cannot overcome the principled ambiguity that there are many possible 3D body pose that can be correctly projected into the corresponding 2D pose.
\noindent
To reduce this ambiguity of 3D lifting from 2D estimates, Pavlakos \etal \cite{pavlakos2018ordinal} use ordinal depth annotation between joint pairs, which is a special case of posebits introduced by Pons-Moll \etal \cite{posebits}.

\noindent
\textbf{Estimating 3D pose using 2D projection information.} Bogo \etal \cite{Bogo:ECCV:2016} fit the 2D keypoints projection of the parametric SMPL \cite{SMPL:2015} body model to 2D predictions from a separate method using an optimization approach. Brau \etal \cite{brau_2016_3d} demonstrated that 2D projection, body pose prior, and body part length information can be used as the training loss objectives for 3D pose prediction. Our method extends the idea of \cite{brau_2016_3d} by introducing additional 3D supervision and paired training on in-the-wild dataset. Kanazawa \etal \cite{kanazawaHMR18} showed that pose and shape parameters of the SMPL body model from monocular images can be learned using a neural network. While their method uses a 2D projection loss of the body model as the main objective, their method also requires an adversarial regularizer against parametric body models. This method can be further improved by using additional labels of 3D pose and SMPL parameters if available. Omran \etal \cite{omran201neural} proposed another deep learning approach to infer the parameters of the SMPL body model, and analyzed performance when varying the input representation (silhouettes, 2D keypoints, part segmentations) and the proportion of 2D and 3D data. Our approach outperforms these methods on several benchmark data sets. 

The above review shows that many methods tackle generalizability on the in-the-wild images using either transfer learning from 2D pose task, or by decoupling the 3D pose estimation into separate 2D keypoint detection and depth regression problems. For methods that decouple the 3D representation \cite{Zhou_2017_ICCV, Yang_3dposeCVPR2018, Dabral:ECCV:2018}, depth information is predicted if 3D labels are available and otherwise some weak supervision constraints (e.g. a parametric body model) are used for regularization. 
In this paper, we propose a new architecture that combines explicit encoding of separate 2D and 3D depth features in hidden space, instead of operating on vectorized 2D predictions as in previous lifting schemes.
Our trained projection network further stabilizes overall 3D prediction accuracy.
\section{Approach} \label{approach}

The method estimates the root (pelvis) relative 3D locations of $K$ human body joints $\mathbf{P} = \{\mathbf{J}_1, \dots, \mathbf{J}_K\}$ in the camera reference frame from a monocular RGB image. Our method assumes that a crop around the subject is available.

A baseline strategy for our goal would be as follows:
Given a training set consisting of pairs of RGB images and their corresponding 3D pose labels $\mathcal{D} = \{ ( \mathbf{I}_n, \mathbf{P}^{GT}_n ) \}_{n=1}^N$, 
we could train a convolution-based neural network $f_{RGB}(\mathbf{I}_n, \theta)$ 
to predict a vectorized representation of 3D joint locations. 
Network parameters $\mathbf{\theta}$ could be trained by minimizing the difference $\mathbf{\mathcal{L}}_{3D}$ between pose prediction and ground truth
\begin{equation} \label{eq:1}
    \mathbf{\mathcal{L}}_{3Dpose} = 
    \frac{1}{N} \sum_{n=1}^{N} \parallel f_{RGB}(\mathbf{I}_n, \theta) - \mathbf{P}^{GT}_n \parallel^2_2
\end{equation}
By training on currently available image data sets with 3D pose annotation, such direct supervision approach can already enable the network to achieve reasonable performance on studio test images. However, such a baseline method is still constrained in its ability to generalize to in-the-wild scenes due to the limited amount of available real world images with ground truth 3D poses.

We therefore introduce several strategies to augment such a 3D pose network such that it performs better on in-the-wild scenes. Our augmented network can be trained on both, images with 3D labels and in-the-wild images with only 2D labels. First, using an explicit 2D pose representation in the feature space of the CNN combined with 2D pre-training can significantly boost the quality of the prediction. Second, 
we propose additional supervision by using a trained projection sub-network that learns weak perspective camera information for projecting 3D pose estimates to the 2D image space.
The overview of our network is shown in Figure \ref{fig:architecture}. 

\subsection{Explicit 2D feature representation for 3D pose prediction} \label{explicit2d}

Martinez \etal \cite{JMartinez:ICCV:2017} showed that a simple neural network is capable of directly regressing 3D human pose with good accuracy by using only vectorized 2D pose as input. This shows that a neural network is able to estimate the structure of natural 3D human pose from corresponding 2D information to some extent.
However, such a lifting scheme can only remedy to some extent the fundamental ambiguity that multiple 3D poses can look the same in 2D.
\cite{pavlakos2018ordinal} showed that additional weak ordinal depth supervision can partially resolve the ambiguity of the problem.

We argue that a 2D-to-3D lifting approach can also be applied on 2D heatmap input instead of the vectorized 2D pose representation. From this observation, we decided to design the convolutional features of our CNN to explicitly encode 2D pose heatmap information. The idea behind this decision is to explicitly decouple 2D pose information from other learned features in the convolutional latent space. The rest of the feature maps can be used by the network to capture other image information related to 3D human pose, such as 3D depth. In this way, the network is guided to learn 3D pose features that are more reliable due to the robust 2D pose prediction and easier to interpret. Furthermore, by using a 2D training loss on this component we allow network to learn useful features from images when 3D pose labels are not available. 

To this end, we design a convolutional feature map $\mathbf{F}_{3D} = [\mathbf{h}_{2D}, \mathbf{d}]$ after the extractor network $f_{RGB}$. This feature map consists of $64$ output channels with a spatial dimension of $16 \times 16$. We use the first $14$ channels to capture the 2D pose information. We optimize this region during training by minimizing the loss compared to the 2D ground truth heatmap in a least square sense. The rest of the feature channels $\mathbf{d}$ are not directly constrained by any explicit loss and will be supervised through the 3D pose, 2D projection, as well as additional pose constraints losses explained later. 

To infer 3D pose from $\mathbf{F}_{3D}$, we first combine explicit 2D heatmaps $\mathbf{h}_{2D}$ and the additional features $\mathbf{d}$ learned by the convolutional encoder by using a simple fully connected layer into a latent vector $\mathbf{z} \in \mathbb{R}^{1024}$. Then, a fully connected network with residual connections $f_{3D}$ is used to learn the vectorized 3D pose representation $\mathbf{p}_{3D}$. We design $f_{3D}$ to be similar to the lifting architecture in~\cite{JMartinez:ICCV:2017}. More specifically, we use a series consisting of four fully connected layers with the width of 1024 and ReLU activations. A residual connection is also incorporated to connect $\mathbf{z}$ with the output of the second layer of $f_{3D}$.

\noindent
\textbf{Bone loss} Several earlier works reported that detection-based approaches using a heatmap or volumetric representation tend to achieve better performance on both 2D and 3D pose estimation tasks than approaches regressing vectorized predictions. However, additional structure aware supervision can lift performance of vectorized prediction to a competitive level~\cite{Sun:ECCV:2017}. Since our method also performs vectorized 3D pose prediction, we complement the 3D training loss $\mathcal{L}_{3Dpose}$ (equation \ref{eq:1}) with a bone supervision loss $\mathcal{L}_{bone}$. For 3D training data, $\mathcal{L}_{bone}$ measures the similarity of the vector between a joint $\mathbf{J}_k$ to its corresponding parent in the kinematic chain to ground truth. For 2D data, it measures the difference of scalar bone lengths to ground truth. 

\subsection{Predicting 2D projection from 3D pose} \label{2d_from_3d}

To further improve our method's ability to utilize 2D pose data for training 3D pose prediction, we train a sub-network to project the predicted 3D pose to the image space. Our camera network $f_c$ predicts the principal coordinate $(c_x, c_y)$ and the focal length $(\alpha_x, \alpha_y)$ parameters of a weak perspective camera model from the given input image. By using the features extracted from the latent representation $\mathbf{z}$, we use a multi-layer perceptron to infer the camera parameters $\mathbf{c} \in \mathbb{R}^4$. During training, a 2D loss $\mathcal{L}_{2Dpose}$ measures the L2 distance between ground truth 2D pose and 2D projection $\mathbf{p}_{2D}$ of the predicted 3D pose:
\begin{equation}
    \mathbf{p}_{2D} = \begin{bmatrix}
    \pi_x(\mathbf{p}_{3D}) \\
    \pi_y(\mathbf{p}_{3D}) 
\end{bmatrix} = \begin{bmatrix}
    \alpha_x  \mathbf{p}_{3D}(x) + c_x \\
    \alpha_y  \mathbf{p}_{3D}(y) + c_y
\end{bmatrix}
\end{equation}

\noindent
Our projection formulation allows the network to learn partial information about the 3D pose even when only 2D pose annotations are available. However, there are no constraints that guarantee the correctness of the predicted depth information. To regularize 3D pose prediction when training on 2D data, we use the additional bone loss $\mathcal{L}_{bone}$ enforcing bone length similarity to ground truth for additional supervision. We randomly pick the bone length of one of the training subjects as the ground truth for every training instance. 

\subsection{Network design} \label{3dpose}
 
We use an adapted ResNet-50~\cite{He2016DeepRL} as the basis of the backbone subnetwork $f_{RGB}$ (Figure~\ref{fig:architecture}) that extracts pose features from 2D images. 
This offers a good trade-off between prediction accuracy and inference time, allowing our network to be optionally used in real-time applications. The original ResNet-50 architecture is used up to level \textit{Res4f} and we train level \textit{Res5a} from scratch without striding while also reducing its number of output channels to $1024$. This extractor network is then followed by the 3D pose regressor network described in \ref{explicit2d}. 

The studio datasets with 3D labels and the outdoor data sets with 2D labels which we train on tend to have slightly differing image statistics due to contrast differences, as well as foreground background augmentations on the 3D data sets. To further mitigate this residual domain gap beyond what our new network architecture can already do by its design, we employ a similar pre-training approach as several earlier 3D pose prediction methods, e.g.~\cite{mono-3dhp2017}.
To this end, we first pre-train our ResNet-50 network on ImageNet features to perform 2D heatmap prediction only. Here, intermediate 2D pose supervision is used on the first $14$ channels of the \textit{res4d} and \textit{res5a} feature maps. The same intermediate supervision is also used later when finetuning the complete network on both 2D and 3D pose data. After pre-training, final training of the full network on both outdoor images with 2D annotations and studio images with 3D annotations results in learned features that generalize well to in-the-wild scenes and yield high accuracy in 3D pose estimation. 

Our algorithm can be modified to handle input images of arbitrary framing around the human, because our subnetwork $f_{RGB}$ is convolutional. For example, we can perform tight bounding box cropping around the detected 2D keypoints before passing the rescaled image into the subsequent sub-network. 


\begin{figure*}
\centering
\begin{subfigure}{.14\textwidth}
  \centering
  \includegraphics[width=.8\linewidth]{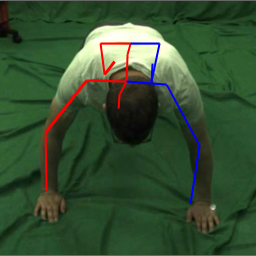}
\end{subfigure}
\begin{subfigure}{.14\textwidth}
  \centering
  \includegraphics[width=.8\linewidth]{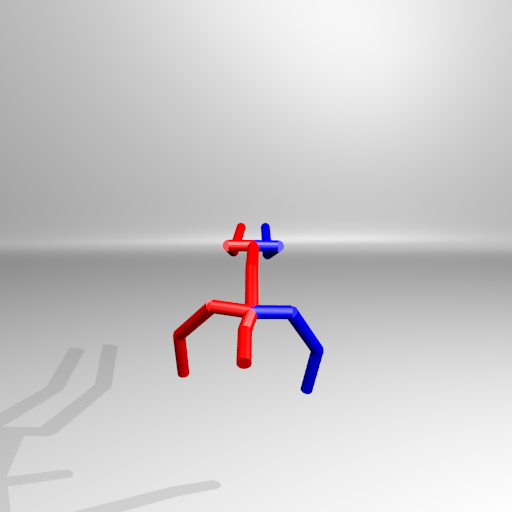}
\end{subfigure}
\begin{subfigure}{.14\textwidth}
  \centering
  \includegraphics[width=.8\linewidth]{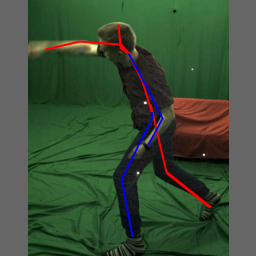}
\end{subfigure}
\begin{subfigure}{.14\textwidth}
  \centering
  \includegraphics[width=.8\linewidth]{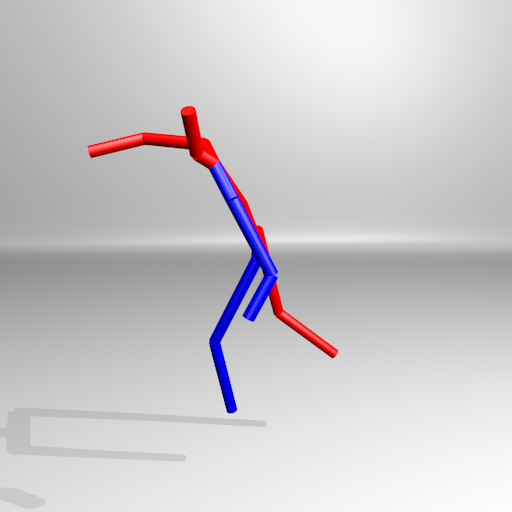}
\end{subfigure}
\begin{subfigure}{.14\textwidth}
  \centering
  \includegraphics[width=.8\linewidth]{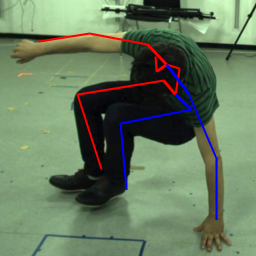}
\end{subfigure}
\begin{subfigure}{.14\textwidth}
  \centering
  \includegraphics[width=.8\linewidth]{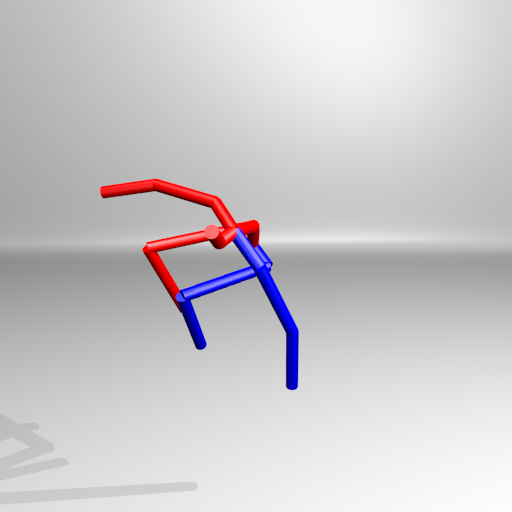}
\end{subfigure}

\vspace{3mm}

\begin{subfigure}{.14\textwidth}
  \centering
  \includegraphics[width=.8\linewidth]{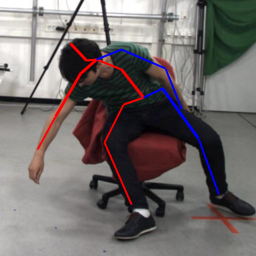}
\end{subfigure}
\begin{subfigure}{.14\textwidth}
  \centering
  \includegraphics[width=.8\linewidth]{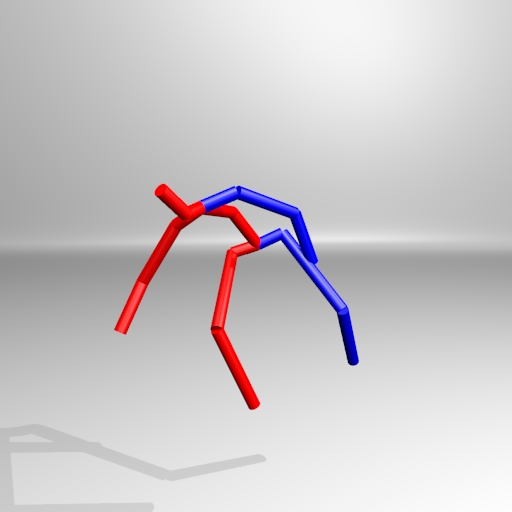}
\end{subfigure}
\begin{subfigure}{.14\textwidth}
  \centering
  \includegraphics[width=.8\linewidth]{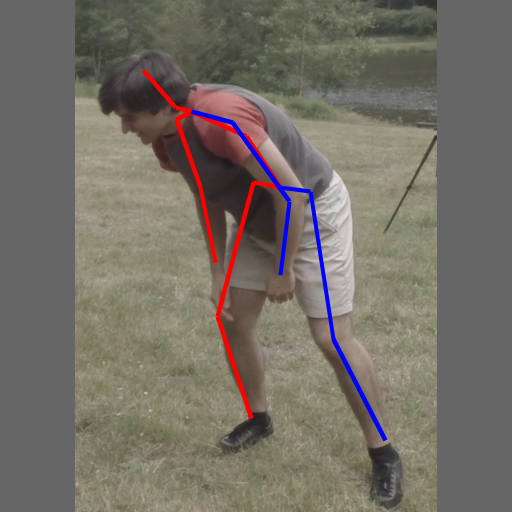}
\end{subfigure}
\begin{subfigure}{.14\textwidth}
  \centering
  \includegraphics[width=.8\linewidth]{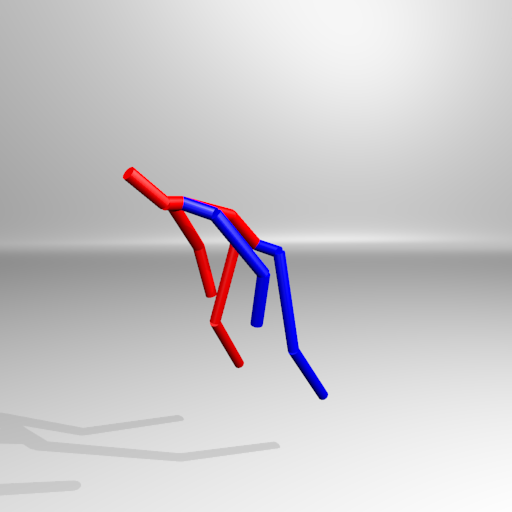}
\end{subfigure}
\begin{subfigure}{.14\textwidth}
  \centering
  \includegraphics[width=.8\linewidth]{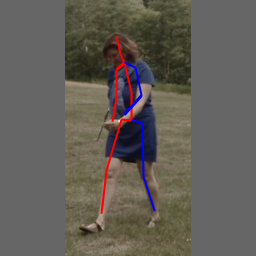}
\end{subfigure}
\begin{subfigure}{.14\textwidth}
  \centering
  \includegraphics[width=.8\linewidth]{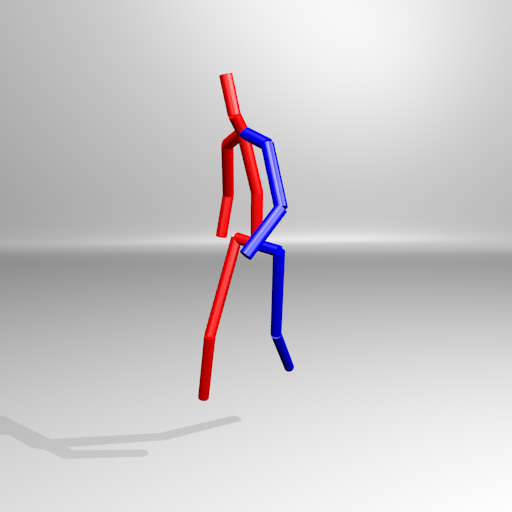}
\end{subfigure}

\vspace{3mm}

\begin{subfigure}{.14\textwidth}
  \centering
  \includegraphics[width=.8\linewidth]{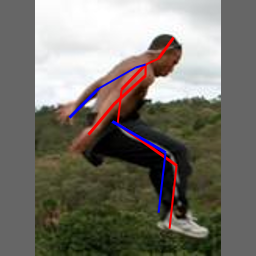}
\end{subfigure}
\begin{subfigure}{.14\textwidth}
  \centering
  \includegraphics[width=.8\linewidth]{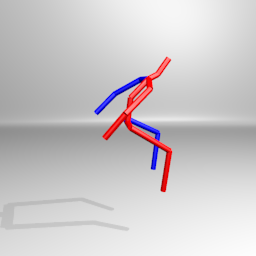}
\end{subfigure}
\begin{subfigure}{.14\textwidth}
  \centering
  \includegraphics[width=.8\linewidth]{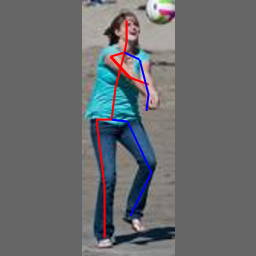}
\end{subfigure}
\begin{subfigure}{.14\textwidth}
  \centering
  \includegraphics[width=.8\linewidth]{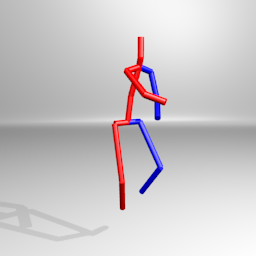}
\end{subfigure}
\begin{subfigure}{.14\textwidth}
  \centering
  \includegraphics[width=.8\linewidth]{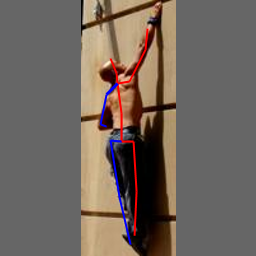}
\end{subfigure}
\begin{subfigure}{.14\textwidth}
  \centering
  \includegraphics[width=.8\linewidth]{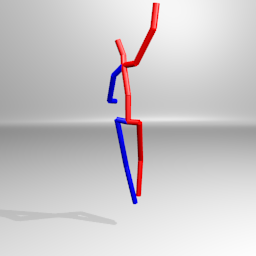}
\end{subfigure}

\vspace{3mm}

\begin{subfigure}{.14\textwidth}
  \centering
  \includegraphics[width=.8\linewidth]{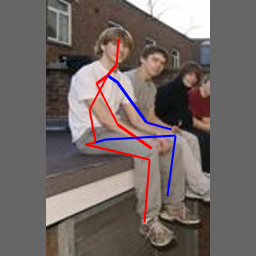}
\end{subfigure}
\begin{subfigure}{.14\textwidth}
  \centering
  \includegraphics[width=.8\linewidth]{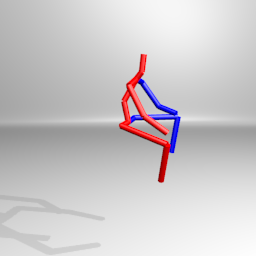}
\end{subfigure}
\begin{subfigure}{.14\textwidth}
  \centering
  \includegraphics[width=.8\linewidth]{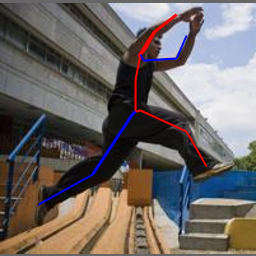}
\end{subfigure}
\begin{subfigure}{.14\textwidth}
  \centering
  \includegraphics[width=.8\linewidth]{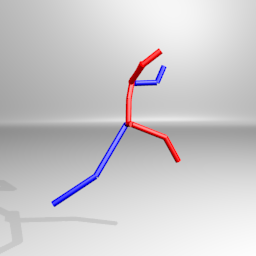}
\end{subfigure}
\begin{subfigure}{.14\textwidth}
  \centering
  \includegraphics[width=.8\linewidth]{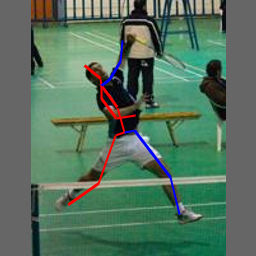}
\end{subfigure}
\begin{subfigure}{.14\textwidth}
  \centering
  \includegraphics[width=.8\linewidth]{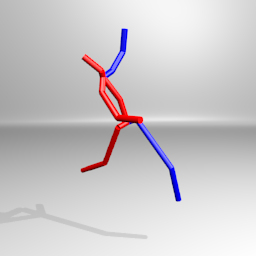}
\end{subfigure}

\caption{Qualitative examples from the MPI-INF-3DHP test set (first to third rows) and LSP (fourth and fifth rows). Refer to the supplementary document for more examples.}
\label{fig:qual_results}
\vspace{-0.3cm}
\end{figure*}

\begin{table}[t]
\small
\centering

\setlength{\tabcolsep}{3pt}
\renewcommand{\arraystretch}{1}

\begin{tabular}{lcccccc}
    \toprule
    \small \textbf{Method}  & \textbf{PCK} & \textbf{PCK} & \textbf{PCK} & \textbf{PCK} & \textbf{AUC} & \textbf{MPJPE} \\
    & \textbf{GS} & \textbf{No GS} & \textbf{Outdoor} & \textbf{All} & \textbf{All} & \textbf{All} \\
    \midrule
    Mehta \cite{mono-3dhp2017} & 84.6  & 72.4  & 69.7 & 76.5 & - & - \\ 
    Mehta \cite{mehta2017vnect} & -  & -  & - & 76.6 & 40.4 & 124.7 \\ 
    Dabral \cite{Dabral:ECCV:2018} & - & - & - & 76.7 & 39.1 & 103.8 \\
    \midrule
   Ours (US) & 87.8 & 80.2 & 73.8 & 81.5 & 44.5 & 90.7 \\
   Ours (GS) & 88.0 & 80.5 & 74.8 & 82.0 & 44.7 & 91.0 \\
   \midrule
   Ours (PA) & 94.9 & 92.4 & 84.0 & 91.3 & 57.5 & 65.4 \\
    \bottomrule
\end{tabular}
\vspace{10pt}

\caption{3D PCK (higher is better) on the MPI-INF-3DHP dataset after training with MPI-INF-3DHP and H3.6M 3D training sets, and MPII and LSP 2D training sets. We clearly outperform all other methods that use a similar combined 2D and 3D training on this benchmark with both indoor and in-the-wild scenes. This holds true for all evaluation protocols (\emph{unscaled} (US), \emph{glob. scaled} (GS), \emph{Procrustes} (PA)).}
\vspace{-0.5cm}
\label{tab:mpii3dhp_result}
\end{table}

\section{Experiments and Discussion} \label{results}

After discussing datasets and network training we will show the high performance of our method qualitatively and quantitatively. 
We use the H3.6M data set~\cite{h36m_pami} to compare general 3D pose estimation accuracy on in-studio data, and show that we outperform previous methods on the more general MPI-INF-3DHP benchmark set. The latter features more diverse motions, and more diverse scenes, including indoor scenes with green screen background ({\bf GS}), as well as more  in-the-wild scenes with general backgrounds, both indoors ({\bf No GS}) and outdoors ({\bf Outdoor}). An ablation analysis shows the significance of the individual components in the proposed approach. The supplementary document contains further explanations and evaluations. 

\subsection{Datasets and evaluation metrics}

As training data with ground truth 3D pose, we use a combination of the H3.6M training set, as well as both background augmented and unaugmented MPI-INF-3DHP training sets which consist of 350k training images in total. As in-the-wild training images with only 2D pose annotation we use the MPII \cite{andriluka14cvpr} and LSP \cite{johnson2012clustered} \cite{johnson2011learning} datasets which are augmented by randomly cropping, translating and rotating the images.

At test time, we compare against other previously proposed methods on both standard H3.6M and MPI-INF-3DHP test data to show both general 3D pose prediction accuracy, as well as state-of-the-art generalization on outdoor scenes. We also qualitatively visualize the state-of-the-art accuracy of our algorithm on in-the-wild images (see Figure~\ref{fig:qual_results}). 

The quantitative performance is evaluated by comparing the Mean Per Joint Position Error (MPJPE), the Percentage of Correct 3D Keypoints (3D PCK) under 150 mm radius~\cite{mono-3dhp2017}, as well as the Area Under Curve (AUC) metric which corresponds to the thresholds of the 3D PCK. Since evaluation protocols in previous work are not uniform, we quantitatively evaluate under the three most commonly used protocols: 
(i) 3D joint predictions are neither scaled nor aligned to ground truth (\emph{unscaled}), (ii) 3D joint predictions are globally scaled with ground truth scale before evaluation (\emph{glob. scaled}), and (iii) 3D joint predictions are aligned to ground truth with full Procrustes alignment (\emph{Procrustes}). We further follow standard practice by cropping a tight bounding box in test images using 2D ground truth information.
Since cropping essentially performs a virtual rotation from the original camera, we use perspective correction \cite{mono-3dhp2017} to re-align the pose to the correct view.

\subsection{Training procedure}

As outlined earlier, we train our network in two stages. We first pre-train the feature extractor network on the 2D heatmap regression task on both MPII \cite{andriluka14cvpr} and LSP \cite{johnson2011learning, johnson2012clustered} datasets. At this stage, the network is trained for 186k iterations with a minibatch size of $21$. The initial learning rate is $0.05$ which we decay exponentially. 

After pre-training, we use the learned weights to initialize the weights of the full 3D pose prediction network. The full network is then trained on both the 3D labeled studio data as well as the in-the-wild data with only 2D annotations. Image data with 3D and 2D annotations are both fed into the network with a minibatch size of $10$ to train for 240k iterations. For this second stage, we again start the training using a learning rate of $0.05$ with a decay over 60k iterations. We use Adadelta with a momentum of $0.9$ in both training stages.

We empirically found that using learning rate discrepancy on the pre-trained layers to preserve in-the-wild features, as suggested by~\cite{mono-3dhp2017}, is necessary to achieve good generalization if 3D training data is very limited or more biased. We found that a learning rate discrepancy with a factor of $100$ when training using H3.6M data as the only source of 3D pose labels yields the best result when tested on the MPI-INF-3DHP dataset. On the other hand, the best performance is achieved without using any such discrepancies when training on both H3.6M and the augmented data of MPI-INF-3DHP as source of 3D labels. This suggest that foreground and background augmentation of the 3D data can further close the domain gap between the indoor and outdoor scenes.

\begin{table*}[t]
\footnotesize
\centering

\setlength{\tabcolsep}{4pt}
\renewcommand{\arraystretch}{1}

\begin{tabular}{lccccccccccccccccc}
    \toprule
    &  \textbf{Direction}  & \textbf{Discussion} & \textbf{Eating} & \textbf{Greeting} & \textbf{Phoning} & \textbf{Posing} & \textbf{Purchases} & \textbf{Sitting} \\
    \midrule
    Mehta* \cite{mono-3dhp2017} & 59.7 & 69.7 & 60.6 & 68.8 & 76.4 & 59.1 & 75.0 & 96.2 \\
    Mehta* \cite{mehta2017vnect} & 62.6 & 78.1 & 63.4 & 72.5 & 88.3 & 63.1 & 74.8 & 106.6 \\
    Pavlakos \cite{pavlakos2017volumetric} & 67.4  & 72.0  & 66.7 & 69.1  & 72.0  & 65.0 & 68.3  & 83.7 \\
    Martinez* \cite{JMartinez:ICCV:2017} & 51.8  & 56.2  & 58.1 & 59.0  & 69.5  & 55.2 & 58.1  & 74.0 \\
    Zhou* \cite{Zhou_2017_ICCV} & 54.8 & 60.7 & 58.2 & 71.4 & 62.0 & 53.8 & 55.9 & 75.2 \\
    Yang* \cite{Yang_3dposeCVPR2018} & 51.5 & 58.9 & 50.4 & 57.0 & 62.1 & 49.8 & 52.7 & 69.2 \\
    Sun* \cite{Sun:ECCV:2017} & 52.8 & 54.8 & 54.2 & 54.3 & 61.8 & 53.1 & 53.6 & 71.7 \\
    Kanazawa* \cite{kanazawaHMR18} & - & - & - & - & - & - & - & - \\
    Luvizon* \cite{luvizon2018cvpr} & 49.2 & 51.6 & 47.6 & 50.5 & 51.8 & 48.5 & 51.7 & 61.5 \\
    Dabral* \cite{Dabral:ECCV:2018} & 46.9 & 53.8 & 47.0 & 52.8 & 56.9 & 45.2 & 48.2 & 68.0 \\
    \midrule
    Ours* (H80K) & 57.1 & 69.6 & 61.6 & 66.0 & 73.4 & 57.1 & 70.9 & 89.8 \\
    Ours* (5 fps) & 54.0	& 65.1 & 58.5 & 62.9 & 67.9 & 54.0 & 60.6 & 82.7 \\
   
   \midrule

    &  \textbf{Sit down} & \textbf{Smoke} & \textbf{Take photo}  & \textbf{Waiting} & \textbf{Walk} & \textbf{Walk dog} & \textbf{Walk pair} & \textbf{Average} \\
    
    \midrule
    Mehta* \cite{mono-3dhp2017} & 122.9 & 70.8 & 85.4 & 68.5 & 54.4 & 82.0 & 59.8 & 74.1 \\
    Mehta* \cite{mehta2017vnect} & 138.7 & 78.8 & 93.8 & 73.9 & 55.8 & 82.0 & 59.6 & 80.5 \\
    Pavlakos \cite{pavlakos2017volumetric} & 96.5  & 71.7  & 77.0 & 65.8  & 59.1  & 74.9 & 63.2  & 71.9 \\
    Martinez* \cite{JMartinez:ICCV:2017} & 94.6  & 62.3  & 78.4 & 59.1  & 49.5  & 65.1 & 52.4  & 62.9 \\
    Zhou* \cite{Zhou_2017_ICCV} & 111.6 & 64.1 & 65.5 & 66.1 & 63.2 & 51.4 & 55.3 & 64.9 \\
    Yang* \cite{Yang_3dposeCVPR2018} & 85.2 & 57.4 & 65.4 & 58.4 & 60.1 & 43.6 & 47.7 & 58.6 \\
    Sun* \cite{Sun:ECCV:2017} & 86.7 & 61.5 & 67.2 & 53.4 & 47.1 & 61.6 & 53.4 & 59.1 \\
    Kanazawa* \cite{kanazawaHMR18} & - & - & - & - & - & - & - & 88.0 \\
    Luvizon* \cite{luvizon2018cvpr} & 70.9 & 53.7 & 60.3 & 48.9 & 44.4 & 57.9 & 48.9 & 53.2 \\
    Dabral* \cite{Dabral:ECCV:2018} & 94.0 & 55.7 & 63.6 & 51.6 & 40.3 & 55.4 & 44.3 & 55.5 \\
    \midrule
    Ours* (H80K) & 109.2 & 68.6 & 81.3 & 65.8 & 54.3 & 78.4 & 58.2 & 71.1 \\
    Ours* (5 fps) & 98.2	& 63.3 & 75.0 & 61.2 & 50.0 & 66.9 & 56.5 & 65.7  \\
    \bottomrule
\end{tabular}

\caption{Mean Per Joint Position Error (MPJPE) on H3.6M when trained on H3.6M (ours are \emph{glob. scaled} for evaluation). \newline 
(*) indicates methods that also use 2D labeled datasets during training or pre-training.}
\vspace{5pt}
\label{tab:h36m_result}
\end{table*}

\begin{table*}[t]
\footnotesize
\centering

\setlength{\tabcolsep}{4pt}
\renewcommand{\arraystretch}{1}

\begin{tabular}{lcccccccccccccccc}
    \toprule
    &  Direct.  & Discuss & Eat & Greet & Phone & Pose & Purch. & Sit & SitD & Smoke & Photo  & Wait & Walk & WalkD & WalkP & Avg.\\
    \midrule
    Sun* \cite{Sun:ECCV:2017} & 42.1 & 44.3 & 45.0 & 45.4 & 51.5 & 43.2 & 41.3 & 59.3 & 73.3 & 51.0 & 53.0 & 44.0 & 38.3 & 48.0 & 44.8 & 48.3 \\
    Kanazawa* \cite{kanazawaHMR18} & - & - & - & - & - & - & - & - & - & - & - & - & - & - & - & 56.8 \\
    Dabral* \cite{Dabral:ECCV:2018} & 32.8 & 36.8 & 42.5 & 38.5 & 42.4 & 35.4 & 34.3 & 53.6 & 66.2 & 46.5 & 49.0 & 34.1 & 30.0 & 42.3 & 39.7 & 42.2 \\
    Omran \cite{omran201neural} & - & - & - & - & - & - & - & - & - & - & - & - & - & - & - & 59.9 \\
    \midrule
   
   Ours* (H80K) & 46.1 & 51.3 & 46.8 & 51.0 & 55.9 & 43.9 & 48.8 & 65.8 & 81.6 & 52.2 & 59.7 & 51.1 & 40.8 & 54.8 & 45.2 & 53.4 \\
   Ours* (5 fps) & 43.7 & 46.9 & 45.4 & 48.0 & 50.2 & 40.6 & 41.6 & 60.7 & 75.6 & 48.8 & 54.9 & 46.8 & 36.9 & 47.5 & 43.9 & 49.2 \\

    \bottomrule
\end{tabular}

\caption{Mean Per Joint Position Error (MPJPE) on H3.6M when trained on H3.6M. (*) indicates methods that also use 2D labeled datasets during training or pre-training. (\emph{Procrustes} for evaluation).}
\label{tab:h36m_result_PA}

\end{table*}

\subsection{Quantitative comparison}

Table~\ref{tab:mpii3dhp_result} compares our method on the MPI-INF-3DHP benchmark against the closest competing approaches that can be trained on both, images with 2D and 3D annotations. All methods were trained using both H3.6M and augmented and unaugmented MPI-INF-3DHP 3D datasets, and the LSP and MPII 2D datasets. Unless stated otherwise, we used the H80K samples of the H3.6M dataset which consists of around 41K training samples before augmentation. Our algorithm achieves by far the highest accuracy (across all evaluation protocols), yielding 82.0$\%$ 3D PCK, 44.7$\%$ AUC and 91.0 mm MPJPE overall (using \emph{glob. scaled} for evaluation). We also achieve the state-of-the-art result specifically on the outdoor scenes with 74.8$\%$ 3D PCK.
Further, the average 3D PCK of 91.3$\%$ is the highest ever reported by all algorithms that evaluated on the MPI-INF-3DHP, irrespective of what training data they used. 
\begin{table}[t]
\small
\centering

\setlength{\tabcolsep}{3pt}
\renewcommand{\arraystretch}{1}

\begin{tabular}{lcccccc}
    \toprule
    \small \textbf{Method}  & \textbf{PCK} & \textbf{AUC} & \textbf{MPJPE} \\
    \midrule
    Mehta \etal \cite{mono-3dhp2017} & 64.7 & 31.7 & - \\
    Yang \etal \cite{Yang_3dposeCVPR2018} & 69.0 & 32.0 & - \\
    Zhou \etal \cite{Zhou_2017_ICCV} & 69.2 & 32.5 & -\\
    \midrule
    Ours (\emph{unscaled}) & 69.6 & 35.5 & 127.0 \\
    Ours (\emph{glob. scaled}) & 70.4 & 36.0 & 129.1 \\
    \midrule
    Ours (\emph{Procrustes}) & 82.9 & 45.4 & 92.0 \\
    \bottomrule
\end{tabular}
\vspace{10pt}

\caption{Comparison on MPI-INF-3DHP after training on H3.6M only. We outperform all other approaches in all metrics and testing protocols.}
\vspace{-0.5cm}
\label{tab:mpii3dhp_train_h36m_result}
\end{table}

Table~\ref{tab:mpii3dhp_train_h36m_result} further shows the comparison of our approach to other methods on MPI-INF-3DHP, when all methods are trained using only H3.6M as the source of 3D pose labels. Also here, our method achieves the highest accuracy in terms of 3D PCK and AUC on the basis of all three evaluation protocols.

\begin{figure*}[ht]
\centering

\begin{subfigure}{.14\textwidth}
  \centering
  \includegraphics[width=.8\linewidth]{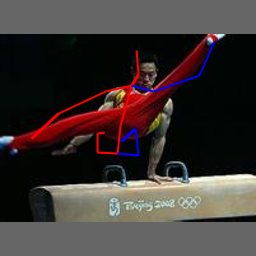}
\end{subfigure}
\begin{subfigure}{.14\textwidth}
  \centering
  \includegraphics[width=.8\linewidth]{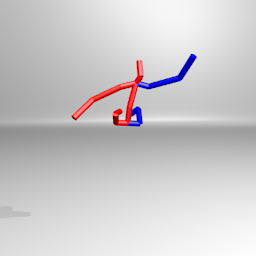}
\end{subfigure}
\begin{subfigure}{.14\textwidth}
  \centering
  \includegraphics[width=.8\linewidth]{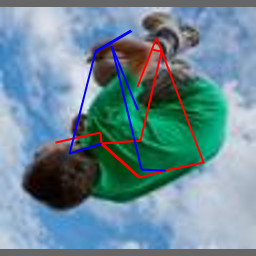}
\end{subfigure}
\begin{subfigure}{.14\textwidth}
  \centering
  \includegraphics[width=.8\linewidth]{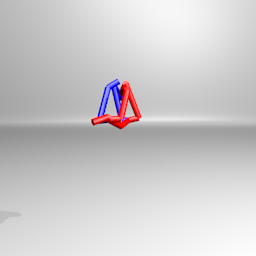}
\end{subfigure}
\begin{subfigure}{.14\textwidth}
  \centering
  \includegraphics[width=.8\linewidth]{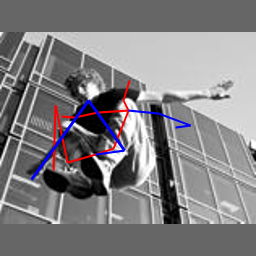}
\end{subfigure}
\begin{subfigure}{.14\textwidth}
  \centering
  \includegraphics[width=.8\linewidth]{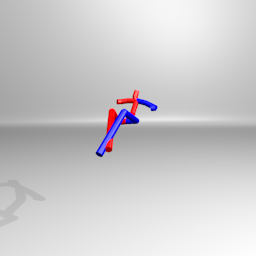}
\end{subfigure}

\vspace{4mm}

\caption{Examples of prediction failures by our proposed method.}
\label{fig:failures}
\vspace{-0.3cm}
\end{figure*}

Finally, we also compare our method by only using the H80K samples of H3.6M as the 3D pose dataset and testing on every 64\textit{th} frame of the S9 and S11 subjects in H3.6M, see Table~\ref{tab:h36m_result} (we use \emph{glob. scaled} following \cite{Zhou_2017_ICCV}\cite{Yang_3dposeCVPR2018}\cite{Dabral:ECCV:2018}) and Table~\ref{tab:h36m_result_PA} (\emph{Procrustes}). On this test set which is heavily biased to in-studio data of a single background our method geared for in-the wild generalization cannot beat the best performing methods. However, it still achieves competitive accuracy. When we increased the number of training data by sampling from H3.6M at 5 frames per second, our method achieved a better MPJPE of 65.7~mm while maintaining competitive result when tested on MPI-INF-3DHP with 71.2$\%$ 3D PCK and 36.3$\%$ AUC. 
When using \emph{Procrustes} during comparison, we achieve a state-of-the-art accuracy of 53.4~mm average MPJPE when trained using H80K samples and 49.2~mm average MPJPE when trained using H3.6M data sampled at 5 fps.
Notably, here we also outperform other methods that use some form of pose projection operation related to our architecture and regularization with a statistical body model, namely~\cite{kanazawaHMR18} and~\cite{omran201neural}. 

\subsection{Ablation study}

We run an ablation study to measure the effectiveness of our proposed contributions (Table \ref{tab:ablation}). We use a direct 3D pose regression method with 2D pose pre-training without the explicit 2D pose loss in the feature space and without the 2D-from-3D projection loss as baseline. The baseline is trained on 3D data only and uses both joint position and bone losses as training objective. We train all of the comparison results on the H80K samples of H3.6M and then performed the evaluation tests on the MPI-INF-3DHP dataset.

The baseline reaches 62.3$\%$ 3D PCK. Using the explicit 2D pose in the latent feature space allows us to use the outdoor data during the training. This addition improves the performance by 3.1$\%$ against the baseline. Similarly, adding the 3D-to-2D projection loss improves the performance of the method even without the explicit 2D pose in latent feature space. Using both the proposed components advances the result to the state-of-the-art result with 70.4$\%$ 3D PCK.

\subsection{Qualitative results and further discussion}

We visualize example prediction results on MPI-INF-3DHP and LSP test images in Figure~\ref{fig:qual_results}. Our method performs consistently well on studio, general indoor and in-the-wild images.

We show several failure cases in Figure \ref{fig:failures}.  Our method can fail on challenging poses which are heavily (self-) occluded, on poses seen from unusual camera angles, or poses which are from what was seen in the training set. Such failure cases are common to many monocular 3D pose estimation approaches. The supplementary document shows additional failure examples of our method.

\section{Conclusion}

We proposed a new deep learning architecture for 3D human pose estimation from  monocular color images. It is designed for training on both, more scarcely available real images with ground truth 3D pose labels, and more widely available in-the-wild images with only 2D pose labels. Our architecture augments a backbone 3D pose inference network with an explicit disentangled 2D pose representation in latent feature space and a learned 3D-to-2D projection model. Our algorithm achieves state-of-the-art performance on the in-studio H3.6M dataset and clearly outperforms related work on the more challenging MPI-INF-3DHP benchmark with in-the-wild images. 

\begin{table}[t]
\small
\centering

\setlength{\tabcolsep}{3pt}
\renewcommand{\arraystretch}{1}

\begin{tabular}{lccccc}
    \toprule
    \small \textbf{Method} & \textbf{PCK} & \textbf{AUC} \\
    \midrule
   Baseline (direct 3D prediction & & \\
   + bone loss) & 62.3 & 30.3 \\
   \hline
   + 2D latent loss & & \\ + outdoor data & 66.4 & 33.0 \\
   \hline
   + 3D-to-2D projection + outdoor data & 69.5 & 35.3  \\
   \hline
   + 2D latent loss + outdoor data + & & \\ 3D-to-2D projection & 70.4 & 36.0  \\
    \bottomrule
\end{tabular}
\vspace{10pt}

\caption{Ablation study on MPI-INF-3DHP test data (split into scene sub-categories: in-studio with green screen (GS), and more in-the-wild scenes indoors (No GS) and outdoors (Outdoor)). 
Only H3.6M data with ground truth 3D labels were used for training. 3D predictions are globally scaled.}
\label{tab:ablation}
\end{table}

\section{Acknowledgments}
\noindent
This work was supported by the ERC Consolidator Grant 4DRepLy (770784). Gerard Pons-Moll is funded by the Deutsche Forschungsgemeinschaft (DFG. German Research Foundation) - 409792180.

{\small
\bibliographystyle{ieee}
\bibliography{article}
}

\clearpage



\section*{Supplementary material}

# Supplemental Document: In the Wild Human Pose Estimation using Explicit 2D Features and Intermediate 3D Representations

## 1. Details of the loss functions used for training

Here we describe the loss functions used to train our neural network.

Given an input image $\mathbf{I} \in \mathbb{R}^{w \times h \times 3}$, the extractor network $f_{RGB}$ will predict the features $\mathbf{F}_{3D}$ which consist of the explicit 2D pose features $\mathbf{h}_{2D}$ and additional pose cues $\mathbf{d}$ as feature maps. The predicted 2D pose features are defined as 2D per-joint heatmaps [2]

$$\mathbf{h}_{2D} = (\mathbf{m}_1, \mathbf{m}_2, \ldots, \mathbf{m}_K),$$
$$\text{where } \mathbf{m}_k \in \mathbb{R}^{\frac{w}{s} \times \frac{h}{s}}.$$

where $s = 16$ is the heatmap down-sampling factor.

Similarly, the ground truth heatmaps are defined as

$$\mathbf{h}^{GT} = (\mathbf{m}_1^{GT}, \mathbf{m}_2^{GT}, \ldots, \mathbf{m}_K^{GT}),$$
$$\text{where } \mathbf{m}_k^{GT} \in \mathbb{R}^{\frac{w}{s} \times \frac{h}{s}}.$$

To train the 2D pose features, we minimize the difference between the predicted 2D joint heatmaps and the ground truth maps using an L2 loss

$$\mathcal{L}_{2Dheatmap} = \sum_{k=1}^{K} b_k \parallel \mathbf{m}_k - \mathbf{m}_k^{GT} \parallel_2^2, \tag{1}$$

where $b_k \in \{0, 1\}$ is a binary mask to ensure that the objective is not evaluated if the annotation of a particular joint is not available.

The latent features $\mathbf{F}_{3D}$ are then used to predict the 3D pose $\mathbf{P}_{3D} \in \mathbb{R}^{21 \times 3}$ by the sub-network $f_{3D}$. Given a 3D pose annotation $\mathbf{P}^{GT} \in \mathbb{R}^{21 \times 3}$, the 3D joint position loss is calculated as follows

$$\mathcal{L}_{3Dpose} = \parallel \mathbf{P}_{3D} - \mathbf{P}^{GT} \parallel_2^2. \tag{2}$$

Given that $Parent(\mathbf{J}_k)$ is the position of the parent of a joint $\mathbf{J}_k \in \mathbb{R}^3$ in the kinematic chain, when the 3D joint position ground truth $\mathbf{J}_k^{GT} \in \mathbb{R}^3$ is available, the bone during training is defined as

$$\mathcal{L}_{bone} = \sum_{k=1}^{K} \parallel \left( Parent(\mathbf{J}_k) - \mathbf{J}_k \right) - \left( Parent(\mathbf{J}_k^{GT}) - \mathbf{J}_k^{GT} \right) \parallel_2^2. \tag{3}$$

On the other hand, if we train on data for which only 2D joint annotations, but no 3D annotations are available, then we instead only compare the bone length magnitude between the predicted joint with a bone length $\mathbf{J}_k^S$ randomly selected from a training annotation

$$\mathcal{L}_{bone} = \sum_{k=1}^{K} \Big\| \parallel Parent(\mathbf{J}_k) - \mathbf{J}_k \parallel_2^2 \\ - \parallel Parent(\mathbf{J}_k^S) - \mathbf{J}_k^S \parallel_2^2 \Big\|_2^2. \tag{4}$$

Finally, given a predicted 2D pose from the projection layer $\mathbf{p}_{2D}$ (Eq. 2 in the main document) and its corresponding ground truth 2D joint coordinates in the image space $\mathbf{p}_{2D}^{GT}$, the projection loss is defined as

$$\mathcal{L}_{2Dpose} = \parallel \mathbf{p}_{2D} - \mathbf{p}_{2D}^{GT} \parallel_2^2. \tag{5}$$

The final training loss can be expressed as

$$\mathcal{L}_{all} = \lambda_{2Dheatmap} \mathcal{L}_{2Dheatmap} + \lambda_{3Dpose} \mathcal{L}_{3Dpose} \\ + \lambda_{bone} \mathcal{L}_{bone} + \lambda_{2Dpose} \mathcal{L}_{2Dpose}, \tag{6}$$

where $\lambda_{3Dpose} = 10$, $\lambda_{2Dheatmap} = 0.1$, $\lambda_{2Dpose} = 10$. $\lambda_{bone} = 10$ if the bone direction is considered (i.e. 3D pose annotations are given) and $\lambda_{bone} = 100$ if we only estimate the bone length scalar (i.e. only 2D annotations are given).

## 2. Additional comparisons on MPI-INF-3DHP

At some point in the past, the authors of MPI-INF-3DHP released a correction to the ground truth annotations of a subset of two their six test sequences. For all our tests, we used the corrected data.

Their very first version of the test set contained small errors on the in-studio sequences with general, i.e. no green screen, background (test subject 3 and 4, meaning sequences labelled **No GS** in the paper and this document).

On these sequences, before correction, the annotations were temporally misaligned by one or two frames.

It is hard for us to say what previous paper we compared against may have unknowingly used the uncorrected subset of sequences.

For our tests to be as transparent and fair as possible, we therefore also provide a comparison on the subset of 4 out of 6 MPI-INF-3DHP test sequences (**GS** and **Outdoors**) that were always correct.

| Method | 3D training data | PCK GS | PCK Outdoor | PCK All | AUC All | MPJPE All |
|---|---|---|---|---|---|---|
| Mehta *et al*. [1] | H3.6M + MPI-INF-3DHP | 84.6 | 69.7 | 78.8 | - | - |
| Mehta *et al*. [1] | H3.6M | 70.8 | 58.5 | 66.0 | - | - |
| Zhou *et al*. [3] | H3.6M | 71.1 | 72.7 | 71.7 | - | - |
| Ours (*unscaled*) | H3.6M + MPI-INF-3DHP | 87.8 | 73.8 | 82.3 | 45.3 | 91.4 |
| Ours (*unscaled*) | H3.6M | 74.6 | 64.0 | 70.5 | 36.3 | 128.7 |
| Ours (*glob. scaled*) | H3.6M (sampled at 5 fps) | 75.4 | 66.9 | 72.1 | 37.2 | 125.5 |
| Ours (*glob. scaled*) | H3.6M + MPI-INF-3DHP | 88.0 | 74.8 | 82.9 | 45.6 | 91.8 |
| Ours (*glob. scaled*) | H3.6M | 75.2 | 65.3 | 71.4 | 36.9 | 131.4 |
| Ours (*glob. scaled*) | H3.6M (sampled at 5 fps) | 75.8 | 67.9 | 72.8 | 37.8 | 128.6 |
| Ours (*Procrustes*) | H3.6M + MPI-INF-3DHP | 94.9 | 84.0 | 90.7 | 58.0 | 66.1 |
| Ours (*Procrustes*) | H3.6M | 85.9 | 78.8 | 83.2 | 46.6 | 91.1 |
| Ours (*Procrustes*) | H3.6M (sampled at 5 fps) | 86.2 | 78.0 | 83.0 | 47.5 | 89.6 |

Table 1: Comparison on the subset of MPI-INF-3DHP test sequences that was not corrected at some point by the authors of MPI-INF-3DHP (GS and Outdoors). All here refers to the average on this subset of sequences. Unless stated otherwise, all H3.6M training data mentioned in this table use H80K samples.

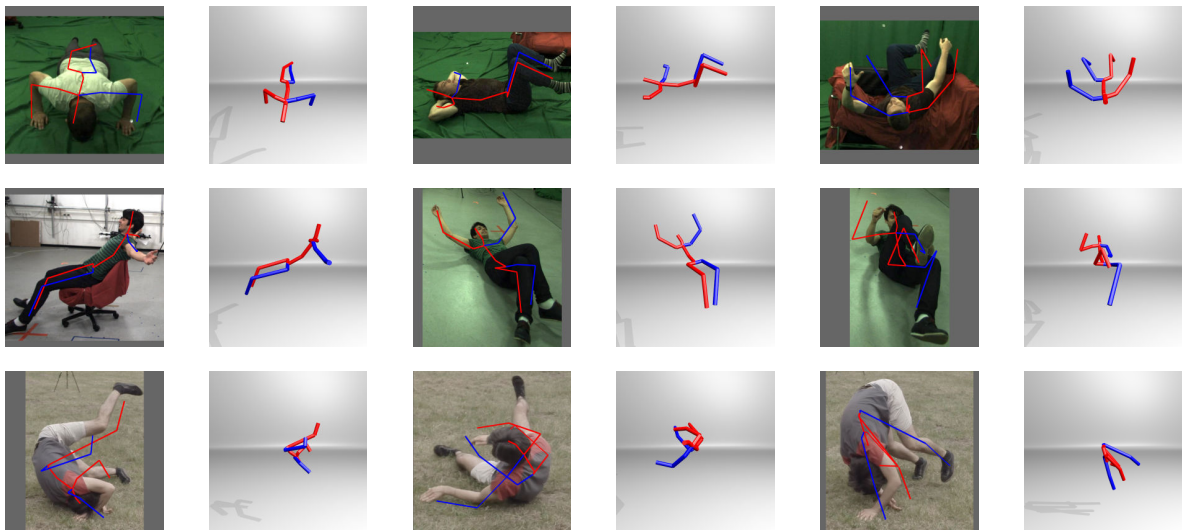

Figure 1: Prediction failure examples by our proposed method on different scenes. Each row from top to bottom represents studio green screen, studio non-green screen, and outdoor scenarios respectively.

Table 1 shows the comparison for methods trained on both H3.6M and MPI-INF-3DHP using the mentioned subset for testing. We include methods that in their original papers reported the respective results on the subsets of test sequences, too. Again, all evaluations of our method in the main paper and here are performed with the corrected annotations. Our proposed method is also state-of-the-art when tested on this subset of sequences.

We also show the activity-wise performance of our method tested on MPI-INF-3DHP in Table 2 respectively. Our method achieves a very high 3D PCK of more than $80\%$ on almost all categories, except for the on-the-floor activities ($60.7\%$), which are in general also challenging for other methods.

## 3. Failure cases

We show additional pose prediction failure cases on different scenes (studio green screen, studio without green screen, outdoor) in Figure 1.

## 4. Additional results

We show additional qualitative results on the MPI-INF-3DHP test set in Figure 2 and the LSP test set in Figure 3. Our approach captures even difficult 3D poses well from a single color image. For more qualitative results please refer to the accompanying video.

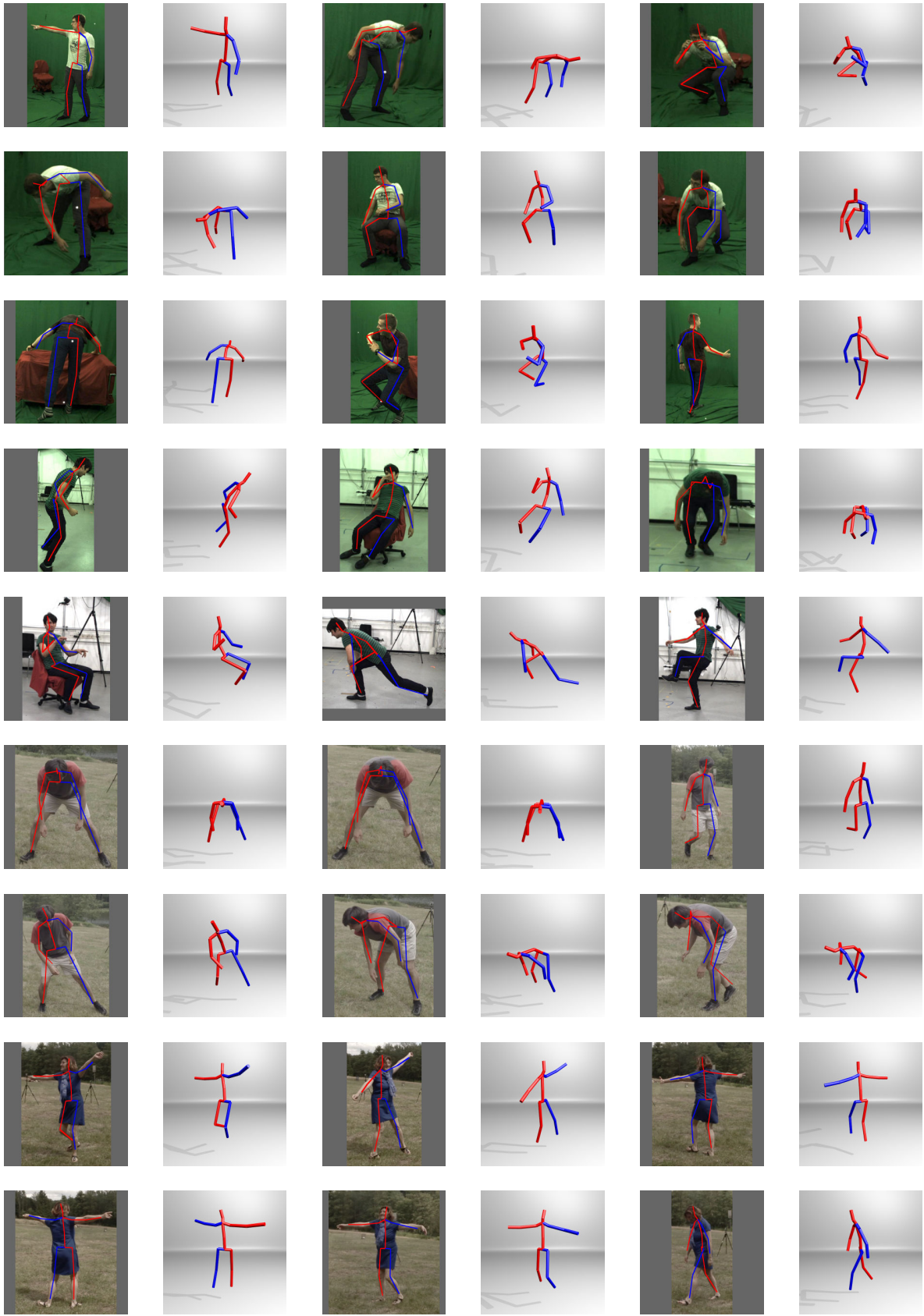

Figure 2: Additional qualitative examples of applying our method to the MPI-INF-3DHP test set.

3

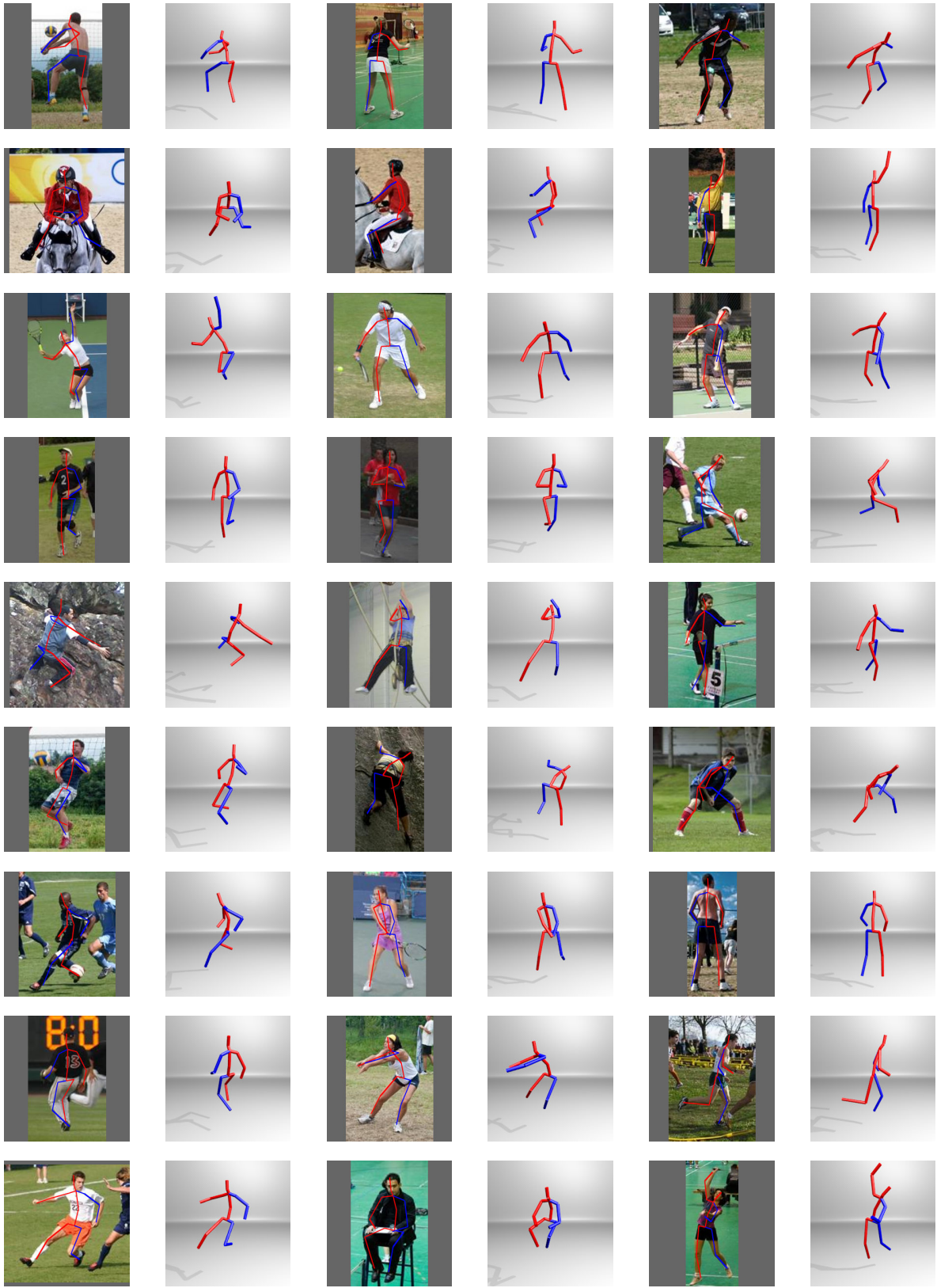

Figure 3: Additional qualitative examples of applying our method on the LSP test set.

4

| Action | PCK | | | | | | | | | AUC |
|---|---|---|---|---|---|---|---|---|---|---|
| | Head | Neck | Shou | Elbow | Wrist | Hip | Knee | Ankle | Total | |
| Standing/Walking | 93.2 | 100.0 | 99.6 | 89.8 | 74.3 | 100.0 | 90.0 | 77.3 | 89.7 | 51.2 |
| Exercising | 91.3 | 98.2 | 98.2 | 87.6 | 75.6 | 100.0 | 77.6 | 65.5 | 85.6 | 47.2 |
| Sitting | 81.7 | 92.8 | 91.8 | 76.7 | 65.1 | 99.8 | 75.8 | 63.9 | 80.0 | 43.7 |
| Reaching/Crouching | 76.6 | 91.1 | 91.3 | 83.3 | 78.0 | 98.7 | 84.2 | 73.2 | 84.6 | 47.6 |
| On The Floor | 62.8 | 83.9 | 78.9 | 54.7 | 40.9 | 94.6 | 53.9 | 28.6 | 60.7 | 28.5 |
| Sports | 90.0 | 99.2 | 98.7 | 84.9 | 67.8 | 100.0 | 90.6 | 72.4 | 87.0 | 49.3 |
| Miscellaneous | 80.8 | 96.8 | 95.3 | 71.3 | 53.8 | 100.0 | 86.5 | 66.9 | 80.4 | 43.4 |
| All | 82.3 | 94.9 | 93.7 | 78.0 | 64.5 | 99.3 | 81.2 | 65.5 | 81.5 | 44.7 |

Table 2: Activity-wise 3D PCK of our method on the MPI-INF-3DHP test set. Our method achieved more than 80% 3D PCK in most actions except for the challenging on-the-floor examples (60.7% 3D PCK).



\end{document}